\documentclass{article}

\makeatletter
\newcommand\tablesize{\@setfontsize\tablesize{8.25}{9.25}}
\makeatother
\PassOptionsToPackage{numbers, compress}{natbib}

\usepackage[preprint]{neurips_2023}

\usepackage[utf8]{inputenc} 
\usepackage[T1]{fontenc}    
\usepackage{hyperref}       
\usepackage{url}            
\usepackage{booktabs}       
\usepackage{amsfonts}       
\usepackage{nicefrac}       
\usepackage{microtype}      
\usepackage{xcolor}         

\usepackage{amssymb}
\usepackage{amsmath}
\usepackage{mathtools}
\usepackage{bm}
\usepackage[ruled,linesnumbered,noend]{algorithm2e}
\usepackage{multirow}
\usepackage{enumerate}
\usepackage{tabularx}
\usepackage[position=top]{subfig}

\DeclareMathOperator*{\argmax}{argmax}
\DeclareMathOperator{\GNN}{GNN}

\DeclareMathOperator{\MLP}{MLP}

\DeclareMathOperator{\IR}{\mathbb{R}}
\DeclareMathOperator{\G}{\mathcal{G}}

\title{Too Big, so Fail? \ -- \ Enabling Neural Construction Methods to Solve Large-Scale Routing Problems }
\author{%
  Jonas K.~Falkner\\
  Department of Computer Science\\
  University of Hildesheim\\ 
  31141 Hildesheim, Germany\\
  \texttt{falkner@ismll.uni-hildesheim.de} \\
  \And
  Lars Schmidt-Thieme\\
  Department of Computer Science\\
  University of Hildesheim\\ 
  31141 Hildesheim, Germany\\
  \texttt{schmidt-thieme@ismll.uni-hildesheim.de}
}

\begin{document}

\maketitle

\begin{abstract}
In recent years new deep learning approaches to solve combinatorial optimization
problems, in particular NP-hard Vehicle Routing Problems (VRP), have been proposed. 
The most impactful of these methods are sequential neural construction
approaches which are usually trained via reinforcement learning. Due to the 
high training costs of these models, they usually are trained on limited
instance sizes (e.g. serving 100 customers) and later applied to vastly
larger instance size (e.g. 2000 customers). By means of a systematic
scale-up study we show that even state-of-the-art neural construction methods are outperformed
by simple heuristics, failing to generalize to larger problem instances.
We propose to use the ruin recreate principle~\cite{schrimpf2000record} 
that alternates between completely destroying a localized part of the solution and 
then recreating an improved variant. 
In this way, neural construction methods like POMO~\cite{kwon2020pomo} are
never applied to the global problem but just in the reconstruction step, 
which only involves partial problems much closer in size to their original training 
instances. 
In thorough experiments on four datasets of varying distributions and modalities 
we show that our neural ruin recreate approach outperforms alternative forms of improving construction
methods such as sampling and beam search
and in several experiments also advanced local search approaches.

\end{abstract}

\section{Introduction}

Neural Construction (NC) methods~\cite{bello2016neural,kool2018attention,joshi2019efficient,falkner2020learning,kwon2020pomo,xin2021multi} have been the driving force behind the success of data driven and machine learning based solution approaches for routing problems since the advent of pointer networks~\cite{vinyals2015pointer} in 2015. 
Apart from the well known Traveling Salesmen Problem (TSP) the most prominent of these problems is the Capacitated Vehicle Routing Problem (CVRP) which involves the planning of several tours to serve a number of $N$ customers from a single depot with capacitated vehicles~\cite{toth2014vehicle}. 
Notwithstanding their success and demonstrated performance on small scale (usually uniform) data, NC approaches exhibit major problems with generalization to larger instances.  
We perform a comprehensive computational study to discover and evaluate the weaknesses of current state-of-the-art NC methods. The results show that generalization to larger problem sizes remains an unsolved problem for all methods no matter the inference strategy. Even advanced and expensive search approaches like Simulation Guided Beam Search (SGBS)~\cite{choo2022simulation} only lead to marginal improvements over a greedy search strategy as soon as instance sizes are more than twice the size of the instances in the training data.
Moreover, existing inference methods fail to make effective use of increased inference times to achieve significant improvements in terms of final performance.

We take these results as motivation to design a new meta control method which transforms the constructive approach into an iterative improvement method applied to smaller sub-graphs of the original problem. The advantages of this formulation are twofold: 
(i) the constructive method is applied to promising sub-graphs with a size close to the training set where NC methods have shown to outperform many heuristic approaches and 
(ii) the iterative formulation can make effective use of additional inference time by focusing on sub-graphs with high potential for further improvement.

\begin{figure}%
	\centering
	\subfloat[\textbf{uniform}]{{\includegraphics[width=0.485\textwidth]{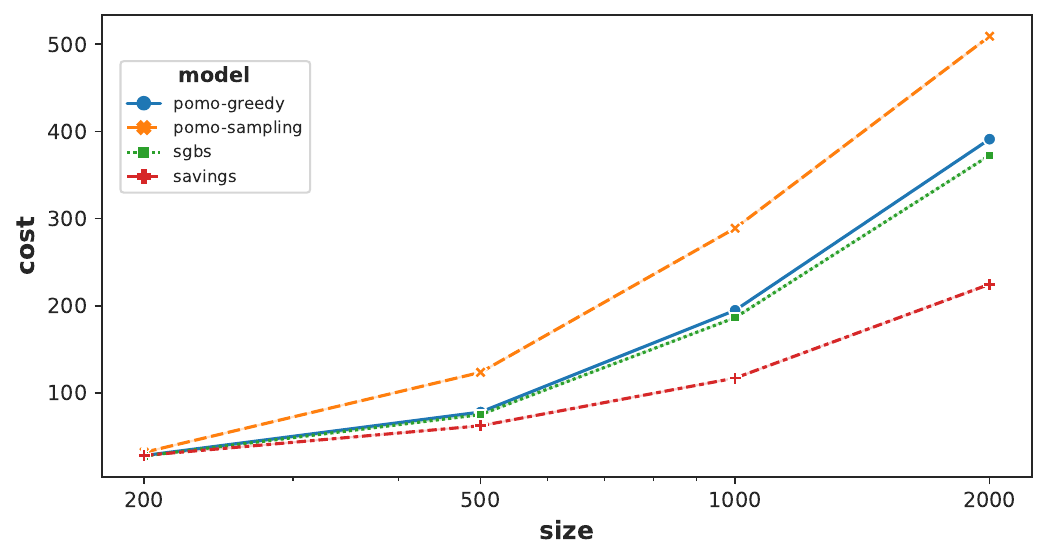} }}%
	\
	\subfloat[\textbf{mixed}]{{\includegraphics[width=0.485\textwidth]{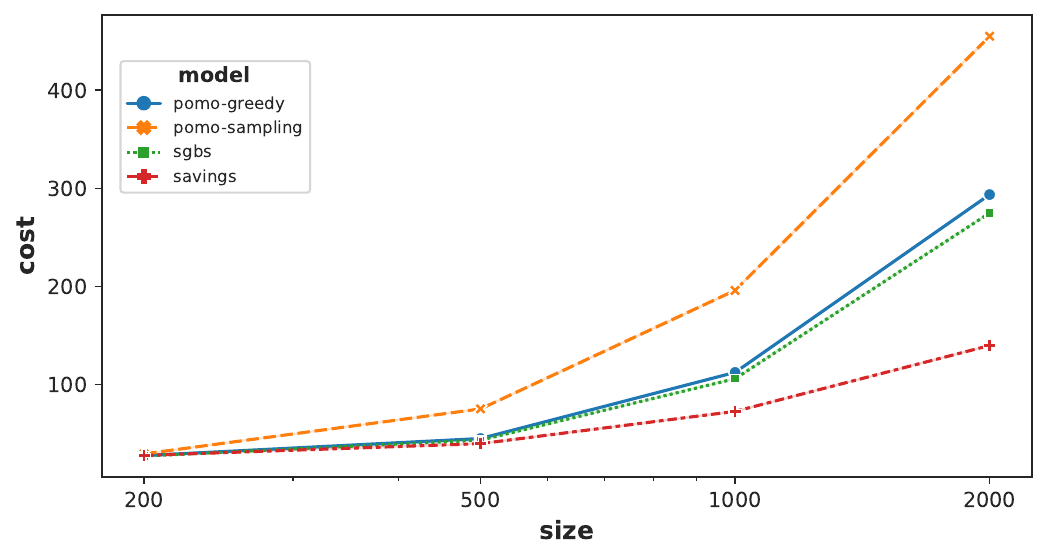} }}%
	\\
	\vspace{-10pt}
	\subfloat[\textbf{Uchoa benchmark}]{{\includegraphics[width=0.485\textwidth]{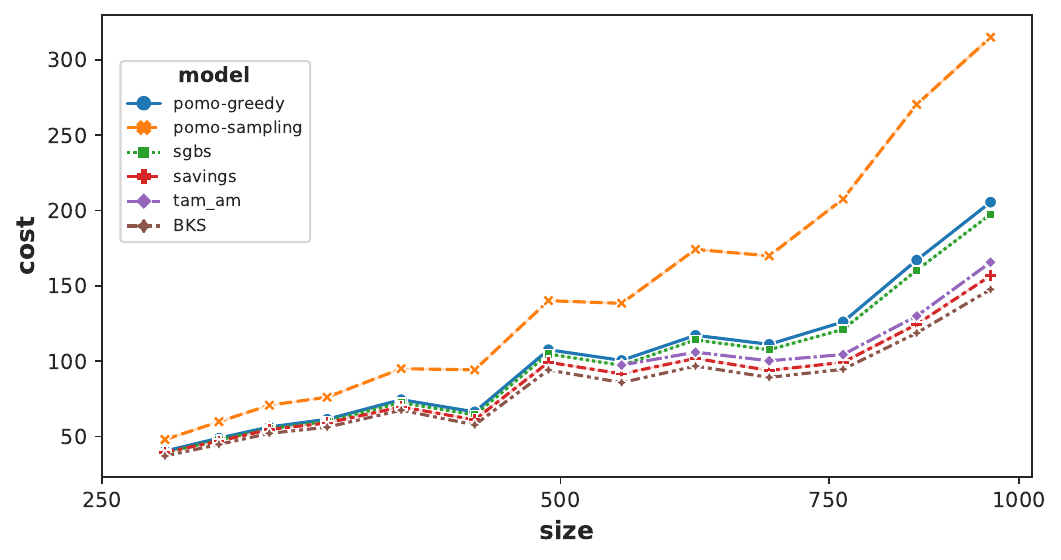} }}%
	\
	\subfloat[\textbf{POMO sampling}]{{\includegraphics[width=0.49\textwidth]{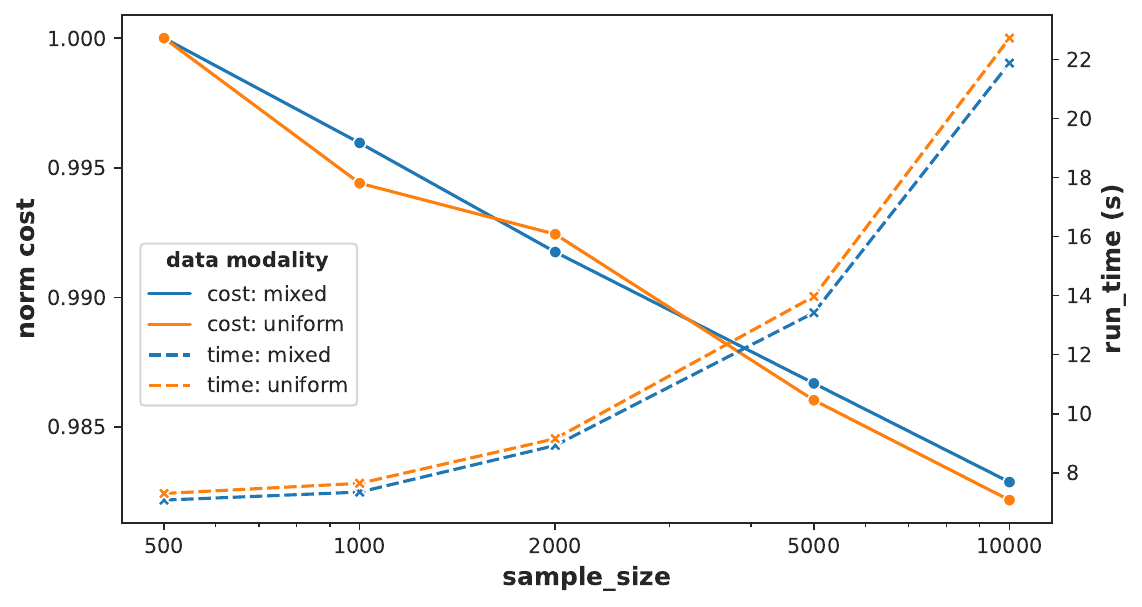} }}%
	\caption{
		\textbf{(a)} and \textbf{(b)}: 
		Results of constructive methods POMO~\cite{kwon2020pomo} (greedy, sampling), SGBS~\cite{choo2022simulation} and Clarke-Wright Savings~\cite{clarke1964scheduling} on uniform and mixed data for different instance sizes. 
		\textbf{(c)}: Results on Uchoa benchmark~\cite{uchoa2017new}. Note that for TAM-AM only results for $N > 500$ were reported in \cite{hou2023generalize}. For the benchmark we also report the \emph{best known solution} (BKS).
		\textbf{(d)}: Normalized cost and run time of the POMO sampling approach for different sampling sizes on uniform and mixed data for instance size $N=500$. 
		Please note the logarithmic scale of the x-axes.
	}%
	\label{fig:constr_methods_comparison}%
\end{figure}

\subsection{Generalization performance of NC methods}\label{ss:nc_generalization}
State-of-the-art constructive methods for the CVRP like POMO~\cite{kwon2020pomo} create a solution sequentially by consecutively adding customer nodes to a tour until the capacity of the vehicle is exhausted. At this time the vehicle returns to the depot and a new tour is started. 
NC methods have shown to outperform simple construction heuristics like the sweep method~\cite{gillett1974heuristic} and the Clarke-Wright savings algorithm~\cite{clarke1964scheduling} on problem instances with uniformly sampled coordinates close to the size $N$ of their training set~\cite{kool2018attention}. 
The sweep method rotates a beam around the depot node adding customer nodes sequentially to a tour in the order they are passed by the beam whereas the Savings algorithm starts with singleton routes and creates tours by consecutively merging the two routes which lead to the largest saving, i.e.\ reduction in total cost.
In Figure~\ref{fig:constr_methods_comparison} we show the performance of POMO and SGBS~\cite{choo2022simulation}, an advanced beam search approach with efficient rollouts and backtracking. 
For POMO we evaluated two different inference strategies, greedy and sampling. The greedy approach performs a greedy rollout by taking the node with maximum probability specified by the learned policy at each step. The rollouts are done in parallel for each possible augmented starting node from which a tour can be created. In contrast, the sampling approach performs a number of rollouts by sampling the next node from the stochastic policy model. 

Figure~\ref{fig:constr_methods_comparison} shows that the models, which were trained on a set of instances of size $N=100$, still achieve reasonable performance for problems of size $N=200$. For these instances of twice the size of the training instances they perform close or on par with the savings algorithm. However, all methods are already significantly outperformed by the heuristic for problems of size $N=500$ and beyond.
In sub-figure~\ref{fig:constr_methods_comparison}(c) we show the results on the well-known Uchoa benchmark~\cite{uchoa2017new}. The data involves varying coordinate, demand and capacity distributions. For that reason and in order to plot a smooth curve to enable a useful comparison, we put the results into bins of size 5 and take the mean over each bin (the non-binned results can be found in the Appendix).
Furthermore, we also show the benchmark results of the recent real-time NC method TAM-AM~\cite{hou2023generalize} and the best known solution as officially reported on CVRPLIB\footnote{\url{http://vrp.galgos.inf.puc-rio.br/index.php/en/}}. As can be seen, even this new state-of-the-art method is outperformed by the savings algorithm on the benchmark.
Similar behavior for NC approaches has been observed for the TSP in~\cite{joshi2021learning}. Even emerging methods like~\cite{bdeir2023attention} which are concerned with improving generalization of NC models to larger instances fail to achieve better performance than the savings algorithm just on problem sizes larger than $N=100$.
While the performance can generally be improved by training on larger instances such that the training distribution is closer to the final test distribution, this is not practical. The reason for that is the significant increase in complexity when training NC methods with reinforcement learning on large instances, since the sequential solution based on policy gradients requires to cache all gradient values for the update. Therefore, in this work we focus explicitly on the \textit{generalization} performance for larger problems.

\subsection{Effective use of available inference time}
Another dimension which has to be considered for NC methods is the actual use of the available inference time. In general, the argument can be made that NC approaches normally should quickly find a reasonable starting solution which then can be improved with heuristics or iterative methods in a second stage. 
However, because of their conceptual simplicity and the easy adaption to new problems, different approaches were proposed to leverage the learned information of the NC policy to find better solutions. These methods usually define some kind of search on the policy model and aim to efficiently traverse the corresponding search space while escaping local optima. The first such approach for NC methods solving routing problems was proposed in~\cite{bello2016neural} and simply samples a number of trajectories from the stochastic policy. This sampling strategy can be seen as a form of scatter search guided by the learned policy but without backtracking. Later methods added smarter starting points, uncorrelated samples, data augmentation and softmax sparsification to the sampling strategy to improve results~\cite{kwon2020pomo,xin2021multi,bdeir2023attention}. Nevertheless, these strategies very quickly have diminishing returns as can be seen in fig.~\ref{fig:constr_methods_comparison}(d). Even the advanced POMO sampling strategy achieves diminishing gains when doubling the sampling size.
Other work proposes more complex search methods utilizing the probabilistic assignment space of the policy model~\cite{hottung2021efficient,kool2022deep,choo2022simulation}. Although such advanced search methods can help to increase the generalization performance, they often incur a significant increase in terms of computational resources and run times while only leading to marginal gains, as can be seen in figure~\ref{fig:constr_methods_comparison} and our experiments in section~\ref{s:experiments}.

Thus, we propose a new method we term \emph{Neural Ruin Recreate} (NRR) designed to enable the use of learned construction heuristics for significantly larger CVRP instances than encountered in the training data.
The main idea is to embed learned NC methods into a powerful improvement procedure which can effectively leverage the learned information of the NC policy model for larger instances.
To that end we combine recent ideas to train a scoring function to estimate the assignment probability of nodes to different subsets~\cite{falkner2023neural} and the expected improvement~\cite{li2021learning} of applying a solution method (which is treated as a black box) to a particular sub problem. 
The resulting scoring function we use to select sub-graphs (SG) of the global solution graph which have a high remaining improvement potential. Our algorithm is defined in terms of a ruin-recreate procedure~\cite{schrimpf2000record} and addresses several detailed design decisions consisting of: 
i) initial solution, 
ii) SG construction, 
iii) SG selection, 
iv) SG solution (update) and
v) acceptance of the update.

\paragraph{Our contributions:}
\begin{enumerate}
	\item We show in a systematic scale-up experiment that neural construction
	methods do not generalize well to problem sizes beyond those seen during
	training. 
	\item We propose a new approach motivated by the well-established ruin recreate
	methodology to enable neural construction methods to efficiently solve routing 
	problems which are up to 40× larger than the instances of their training set.
	\item In a rigorous comparative study we evaluate the efficacy 
	of our method compared to state-of-the-art constructive and improvement 
	approaches. Our method shows competitive results significantly 
	outperforming all other methods on the most difficult instances of the 
	Uchoa benchmark~\cite{uchoa2017new} and the real-world instances of~\cite{li2021learning}. We release all our models and code\footnote{\url{https://github.com/jokofa/NRR}}.
\end{enumerate}

\section{Preliminaries}
\textbf{Problem Formulation} \ 
The capacitated vehicle routing problem (CVRP) is an important NP-hard combinatorial optimization problem~\cite{toth2014vehicle}. It extends the well-known Traveling Salesmen Problem (TSP) to the case with multiple vehicles. It is concerned with serving a number of $N$ customers with coordinates in $\IR^2$ from a single depot. Each customer $n$ has a demand $q_n > 0$ that needs to be served by $K$ vehicles with homogeneous capacities $Q$. Moreover, every tour has to start and end at the depot node and every customer node has to be visited exactly once. The objective is to minimize the total length of all tours in terms of a distance measure $\delta: \IR^2 \to \IR$ (usually euclidean distance).

\textbf{Neural Construction Methods} \ 
Neural construction (NC) methods~\cite{bello2016neural,kool2018attention,joshi2019efficient,falkner2020learning,kwon2020pomo,xin2021multi} create a solution sequentially one node at a time. They normally utilize an encoder-decoder model where the encoder embeds the current problem state and the decoder is queried at each step to select the next node to add to the currently constructed tour. If the depot node is selected, the vehicle returns and a new tour is started. 
A masking scheme ensures that the problem constraints are satisfied. In the simplest case each customer which has already been visited is masked as well as any customer with a demand $q$ larger than the remaining capacity $Q_k$ of the currently employed vehicle $k$.
However, to tackle more complex problems advanced masking schemes can be employed~\cite{falkner2020learning,li2021heterogeneous}.

\subsection{Ruin Recreate Principle}\label{ss:rr}
{
	\SetKwComment{Comment}{// }{ }
	\SetKwFunction{init}{init}
	\SetKwFunction{select}{select\_region}
	\SetKwFunction{ruin}{ruin}
	\SetKwFunction{rec}{recreate}
	\SetKwFunction{acpt}{accept}
	\SetKwInOut{inp}{input}
	\DontPrintSemicolon
	\begin{algorithm}[b]
		\caption{Ruin Recreate}\label{alg:rr}
		
		\inp{Solution space $S$, cost function $c$, stopping criterion} 
		\BlankLine
		$s \gets \init(S)$	\Comment*[r]{Construct initial solution}
		\While{not stopping criterion}{	
			$g \gets \select(s)$ 		\Comment*[r]{Choose part of solution to ruin}
			$s' \gets \ruin(s, g)$	\\
			$s' \gets \rec(S, s')$	\\ 
			$s \gets \acpt(s, s', c)$	\Comment*[r]{Decide wether to accept new solution}
		}
		\Return{s}
	\end{algorithm}
}
The ruin recreate (RR) principle was coined by~\cite{schrimpf2000record} and is a general approach to solve complex combinatorial optimization problems. The key idea of RR is to first \emph{ruin}, i.e.\ destroy, a significant part of the current problem solution and then to \emph{recreate} the destroyed part leading to a better overall solution. This concept is in strong contrast to local search (LS) approaches~\cite{aarts2003local} which are commonly used to optimize routing problems and usually apply small changes to a solution to end up at a close but slightly different solution. In \cite{schrimpf2000record} it was demonstrated that, in particular for complex or very large problems, the RR approach leads to significantly better solutions than comparable LS methods, since it is able to better navigate the corresponding rough search spaces and to more easily escape from local optima by doing "large steps". On top of this procedure often an acceptance method like \textit{Threshold Accepting}~\cite{dueck1990threshold} or \textit{Simulated Annealing} (SA)~\cite{kirkpatrick1983optimization} is used to guide the underlying search. The general procedure is described in Algorithm \ref{alg:rr}. 
For routing problems, the ruin step usually consists of removing a subset of the edges in a particular region of the global solution graph. Then the recreation step is concerned with reinserting useful edges to recreate a feasible solution.

The RR concept has been reinvented several times over the years and is known under different names. To the best of our knowledge the principle was first applied in the method proposed in~\cite{dees1981performance} for wiring of electronic circuits, which was called \textit{Rip-Up and Reroute}. Later, \textit{Large Neighborhood Search} (LNS)~\cite{shaw1998using}, \textit{Ruin Recreate} (RR)~\cite{schrimpf2000record} and \textit{Partial OPtimization Metaheuristic Under Special Intensification Conditions} (POPMUSIC)~\cite{ribeiro2002popmusic} were introduced, which apply the same principle to optimize VRPs.
In the machine learning (ML) field the concept was first used with some learned components by \textit{Neural Large Neighborhood Search} (NLNS)~\cite{hottung2019neural} and \textit{Learning to Delegate} (L2D)~\cite{li2021learning}.
However, all of these related approaches use slightly different strategies for the region selection, ruin and recreate steps (Alg.~\ref{alg:rr} lines 3, 4 and 5 respectively). Regarding the ML approaches NLNS employs an NC method in the recreate step to repair randomly destroyed sub-solutions, whereas L2D utilizes a learned method for region selection while the recreation is done with heuristic solvers. Thus, we are the first to propose a combined neural RR procedure using learned methods for both, selection (neural scoring function) as well as recreate (NC method) steps. %
Moreover, we are the first to draw attention to the existence of the different RR methods which implement similar concepts under varying names and do the first comprehensive comparison of the corresponding approaches (see section \ref{s:experiments}).

\section{Proposed Method}
\begin{figure}[t]
	\centering
	\includegraphics[width=0.8\textwidth]{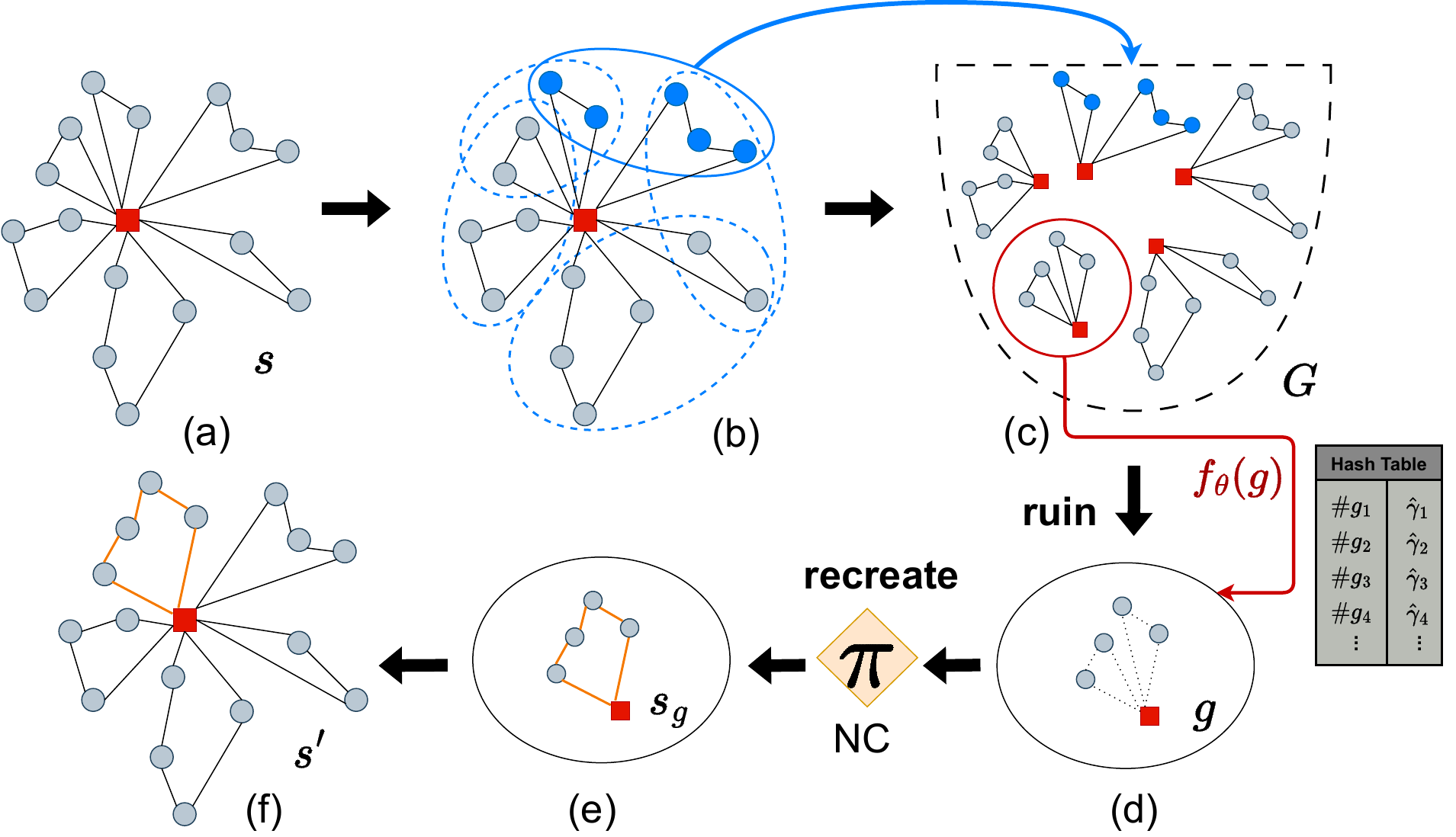}
	\caption{
		One iteration of our NRR procedure: 
		from the current solution graph $s$ \textbf{(a)} several SGs are constructed \textbf{(b)} and put into the set $G$ \textbf{(c)}. Then a promising SG $g$ is selected according to its improvement potential $\hat{\gamma}_g$ estimated by the neural scoring function $f_{\theta}$ and its edges are removed \textbf{(d)}. Finally, the NC method $\pi$ is applied to recreate a new sub-graph solution $s_g$ \textbf{(e)} which then is inserted into the global solution graph $s$ to arrive at the new candidate solution $s'$ \textbf{(f)}.
	} 
	\label{fig:nrr_procedure}
\end{figure}

In order to apply NC methods to much larger problem sizes we propose to combine them with a ruin recreate type of improvement approach. The motivation is to apply the fully trained NC method to problems with a size close to that of their respective training instances. Our procedure first creates a set $G$ of sub-graphs from the global graph $s$ representing the current solution. Then we aim to select a suitable sub-graph $g$ from the set $G$. This SG is ruined by completely removing all of its edges (which represented the original part of the solution). Finally, we recreate a new solution $s_g$ for sub-graph $g$ with the respective NC method $\pi$ and then re-insert it into the global solution $s$.
We describe the method in Algorithm~\ref{alg:nrr} and visualize the general procedure in Figure \ref{fig:nrr_procedure}.
In the following sections we describe the different building blocks which constitute our method.

\subsection{Initial solution}
Usually NC methods without advanced search procedures are concerned with constructing an initial solution. These initial solutions are crucial for following iterative improvement algorithms to find good final solutions. In particular, existing work~\cite{chen2019learning,wu2021learning} has shown that randomly constructed initial solutions can often lead to sub-optimal improvement results when used for procedures which do not require significant levels of randomness as essential part of their improvement strategy (e.g.\ LKH3~\cite{helsgaun2000effective} requires random starts for each of its trials).
However, since the studied NC methods fail to produce decent initial solutions for large-scale problems, we use the Clarke-Wright savings algorithm~\cite{clarke1964scheduling} which showed strong results in our preliminary generalization study (see section \ref{ss:nc_generalization}). 

\subsection{SG construction}
Another crucial component of our algorithm is the construction of the set of sub-graphs $G_t$ from the global solution graph $s_t$ at iteration $t$ shown in Figure~\ref{fig:nrr_procedure} (b) and (c) and Alg.~\ref{alg:nrr} (line 3). In the following we will omit the iteration index $t$ to simplify notation. The set $G$ represents the available SGs which can be selected for further improvement. The key idea here is to put sub-graphs $g$ into $G$ which are of approximately the same size $N_g \approx N_{\text{train}}$ as the training instances of the respective NC method. 
There are different possible approaches one can utilize to construct suitable SGs. We choose to use the tours in the solution graph $r \in s$  as an intermediate graph structure to facilitate the efficient construction of SGs. Furthermore, this approach has the direct advantage of grouping relevant graph parts on a local scale, a concept which is used in many combinatorial optimization methods~\cite{taillard1993parallel,rochat1995probabilistic}. The insight is that optimal tours for the CVRP usually consist of nodes which are in each others local vicinity, while far away nodes normally do not belong to the same tour. 
The selection of different tours to create a respective sub-graph is facilitated by representing each tour by the geometric center $ \mu_r = \frac{1}{|r|} \sum_{n \in r} x_n$ of the coordinates $x_n \in \IR^2$ of all nodes $n \in r$, where $|r|$ is the size of tour $r$, i.e.\ the total number of customers belonging to that tour. 
We experimented with different heuristics to construct suitable SGs from these tours. Details on the exact procedure can be found in the Appendix.

{
	\SetKwComment{Comment}{// }{ }
	\SetKwFunction{init}{init}
	\SetKwFunction{constr}{construct}
	\SetKwFunction{select}{select}
	\SetKwFunction{drop}{drop\_edges}
	\SetKwFunction{insert}{insert}
	\SetKwFunction{acpt}{accept}
	\SetKwInOut{inp}{input}
	\DontPrintSemicolon
	\begin{algorithm}[t]
		\caption{Neural Ruin Recreate}\label{alg:nrr}
		
		\inp{
			Solution space $S$, cost function $c$, stopping criterion, 
			trained NC method $\pi$
		} 
		\BlankLine
		$s \gets \init(S)$				\Comment*[r]{Construct initial solution}
		\While{not stopping criterion}{	
			$G \gets \constr(s)$		\Comment*[r]{construct set of SGs}
			$g \gets \select(G)$		\Comment*[r]{select SG}
			$g \gets \drop(g)$			\Comment*[r]{ruin SG by removing all edges}
			$s_g \gets \pi(g)$			\Comment*[r]{recreate SG solution via NC}
			$s' \gets \insert(s, s_g)$	\Comment*[r]{insert SG into solution}
			$s \gets \acpt(s, s', c)$	\Comment*[r]{Decide wether to accept new solution}
		}
		\Return{s}
	\end{algorithm}
}

\subsection{SG selection}
The next step of our algorithm (Alg.~\ref{alg:nrr}, line 4) is concerned with the selection of SG $g \in G$ (in case of disjoint optimization several $g_1, \dots, g_k \in G$) to be ruined and recreated in the following step. 
In order to select useful SGs with a high remaining improvement potential, we follow \cite{li2021learning,falkner2023neural} in learning a neural scoring function $f: \G \to \IR$. 
This scoring function takes a sub-graph $g$ as input and assigns it a scalar score $\hat{\gamma}_g$ signifying the remaining potential for improvement when applying the respective NC method to recreate its solution. 
It is trained via regression on the actual improvement $\gamma_g$ achieved by the NC method on a large set $D$ of sub-graphs to estimate the remaining potential for unseen SGs during inference. The training set is created by running the respective NC method on varying sub-graphs (whose edges were removed) produced by the prior construction step and registering the achieved improvement $\gamma_g$ as regression target for SG $g$.
Then we can learn a model to represent $f_{\theta}$ with parameters $\theta$ by minimizing the mean squared error (MSE) according to
\begin{equation}
	\mathcal{L}_{MSE} = \frac{1}{|D|} \sum_{i=1}^{|D|}(\gamma_g - f_{\theta}(g))^2.
\end{equation}
In order to fully capture the SG structure for accurate estimates, similar to \cite{falkner2023neural} we use graph neural networks (GNNs)~\cite{morris2019weisfeiler} to encode all nodes as latent vectors $\omega_{\text{node}} \in \IR^{d_{\text{emb}}}$ of dimension $d_{\text{emb}}$:
\begin{equation}
	\omega^{(l)}_i = 
	\GNN^{(l)}(\omega^{(l-1)}_i) =
	\sigma\left( \MLP^{(l)}_1(\omega^{(l-1)}_i) + \MLP^{(l)}_2(\sum_{j\in \mathcal{H}(i)} e_{ji} \cdot \omega^{(l-1)}_j) \right),
\end{equation}
where $\omega^{(l-1)}_i \in \IR^{1 \times d_{\text{emb}}}$ represents the embedding of node $i$ at the previous layer $l-1$, $\mathcal{H}(i)$ is the 1-hop graph neighborhood of node $i$, $e_{ji}$ is the directed edge connecting nodes $j$ and $i$, $\MLP_1$ and $\MLP_2$ are Multi-Layer Perceptrons $\MLP: \IR^{d_{\text{emb}}} \to \IR^{d_{\text{emb}}}$ and $\sigma()$ is a suitable activation function. In order to capture more global information, which is essential for accurate predictions, the node neighborhoods $\mathcal{H}(i)$ are based on the fully connected representation of the original problem graph and consist of the $k$ nearest neighbors of each node in terms of euclidean distance. The input features to the encoder model are the coordinates $x_n$ and demands $q_n$ of each customer node $n \in s$.

Next, pooling is used to aggregate the node information for each SG by summing over the node dimension, producing sub-graph embeddings $\omega_g$. Moreover, we pool over all node embeddings after each GNN layer and aggregate them to create a complete embedding of the solution graph $\omega_s$ which is concatenated with the SG embeddings $\omega_g$ and fed into a final regression head consisting of a stack of MLP layers. Furthermore, we use ReLU~\cite{nair2010rectified} and layer normalization~\cite{ba2016layer}. 
To better understand the effect of the chosen model architecture and pooling operators we perform an ablation study and report the results in the Appendix. Interestingly our model based on the simple order-1 GNN proposed by \cite{morris2019weisfeiler} significantly outperforms the more sophisticated graph attention networks (GATs)~\cite{velivckovic2017graph} which were also used in \cite{li2021learning}.
Since the update of a particular SG leaves the rest of $s$ unchanged, it is very likely that a large subset of the same SGs is encountered in $G$ for several iterations. Hence, we can achieve efficient processing by caching the scores for all evaluated SGs in a hash table for fast lookup, allowing to skip SGs which were already encountered in earlier iterations.
After the score assignment different strategies can be utilized for the final selection of sub-graphs $g$. The simplest option is a \textit{greedy} approach which always selects the SG with the highest score: 
$g = \argmax(\Gamma_G)$ where $\Gamma_G$ is the set of scores $\gamma_g\ \forall g \in G$. Alternatively we may apply the softmax function 
to $\Gamma_G$ and treat the set of scores as a distribution, \textit{sampling} according to the resulting probabilities. 
Finally, we can sample a subset of \textit{disjoint} SGs, optimize all of them and reinsert the ones into $s$ which led to an improvement.
In preliminary experiments we found that the last approach leads to the best results.

\subsection{SG solution}
The solution of the SG is the main part of the ruin recreate procedure. 
First, the SG is ruined by completely dropping all of its edges (see Fig.~\ref{fig:nrr_procedure} (d), Alg.~\ref{alg:nrr}, line 5). Thereby all existing tours in the SG are destroyed and all customer nodes are marked as not being served. This is in line with the general RR principle which requires the complete destruction of a substantial part of the solution~\cite{schrimpf2000record}. 
After the ruin step the SG is treated as an independent CVRP and fed into the NC method $\pi$ to recreate it by constructing a new solution (Fig.~\ref{fig:nrr_procedure} (e), Alg.~\ref{alg:nrr}, line 6).
A suitable configuration for the NC method has to be chosen to achieve useful improvement of $g$ within reasonable time to be able to execute a sufficient number of iterations. For POMO the trade-off is mostly between a greedy decoding strategy and sampling with different number of samples. In general, we found that using SGBS in the recreate step requires too much time, limiting the number of iterations and in turn hurting final performance. 

\subsection{Acceptance}
The canonical RR procedure includes an explicit acceptance strategy for updates to increase the likelihood of escaping local optima. Hence, we employ the well-known simulated annealing (SA)~\cite{kirkpatrick1983optimization} approach to control the acceptance of altered candidate solutions $s'$ which are created through the insertion of the recreated sub-graph $g$ into the previous solution graph $s$.

\section{Related Work}

Construction methods are one of the most prominent ML methods to solve routing problems. First, PointerNetworks~\cite{vinyals2015pointer} were introduced which use an encoder-decoder model with a masking mechanism called \textit{pointer attention}. In following works this approach was improved by adding RL-based training~\cite{bello2016neural}, extending it to VRPs~\cite{nazari2018reinforcement}, replacing RNNs with Transformers~\cite{kool2018attention}, stabilizing training via multiple starting points POMO~\cite{kwon2020pomo} or mutliple decoders MDAM~\cite{xin2021multi} and adding advanced search strategies via simulation SGBS~\cite{choo2022simulation}, dynamic programming~\cite{kool2022deep} or active search~\cite{hottung2021efficient}. 
TAM-AM~\cite{hou2023generalize} employs a different approach learning a model to partition a VRP into subsets where each subset corresponds to a TSP which satisfies the global constraints and then using an NC model to solve these TSP instances independently from the global CVRP. 
Apart from NC methods also neural improvement approaches have been proposed which start at a solution and then iteratively improve it.
Wu et al.~\cite{wu2021learning} propose to parameterize heuristic operators like 2-opt with a learned policy model, an approach further improved in~\cite{ma2021learning} with Dual-Aspect Collaborative Transformers (DACT), while a meta-controller for local search is learned in~\cite{falkner22learning}.
NeuRewriter~\cite{chen2019learning} selects a region and rewrites it based on a heuristic rule set. LCP~\cite{kim2021learning} improves this approach by repeatedly rewriting several segments in parallel. 
NLNS~\cite{hottung2019neural} employs an NC method as repair operator in a Large Neighborhood Search (LNS)~\cite{shaw1998using} whereas a hierarchical problem decomposition approach is proposed in~\cite{zong2022rbg}.
While existing methods utilize the NC model to repair partial solutions, we use them on a much larger scale, employing them to fully recreate a complete subgraph (which is a substantial part of the global solution graph) from scratch.
Learning to delegate (L2D)~\cite{li2021learning} selects regions to update based on the estimated improvement but use heuristics to reconstruct the chosen sub-solutions. In contrast, we reformulate the procedure in a more principled and general way by drawing the connection to the well established RR principle and employ NC methods to recreate SG solutions. Our results show that the careful design of the algorithm leads to significant performance increases. 
A different approach is used for NeuroLKH~\cite{xin2021neurolkh} which employs a learned model for the prior selection of the edge candidate set for the well-known heuristic solver LKH~\cite{helsgaun2000effective}, which is commonly used as a baseline to compare ML based routing algorithms.
Moreover, some prior work regarding generalization performance has been done by Jian et al.~\cite{jiang2022learning} who use meta-learning to generalize to different coordinate distributions while Fu et al.~\cite{fu2021generalize} devise a method to generalize a smaller pre-trained model to larger TSP instances, but require the expensive solution of a set covering problem.
A detailed overview is given in \cite{bengio2021machine,mazyavkina2021reinforcement}.


\begin{figure}
	\centering
	\begin{minipage}{.49\textwidth}
		\centering
		\includegraphics[width=.999\linewidth]{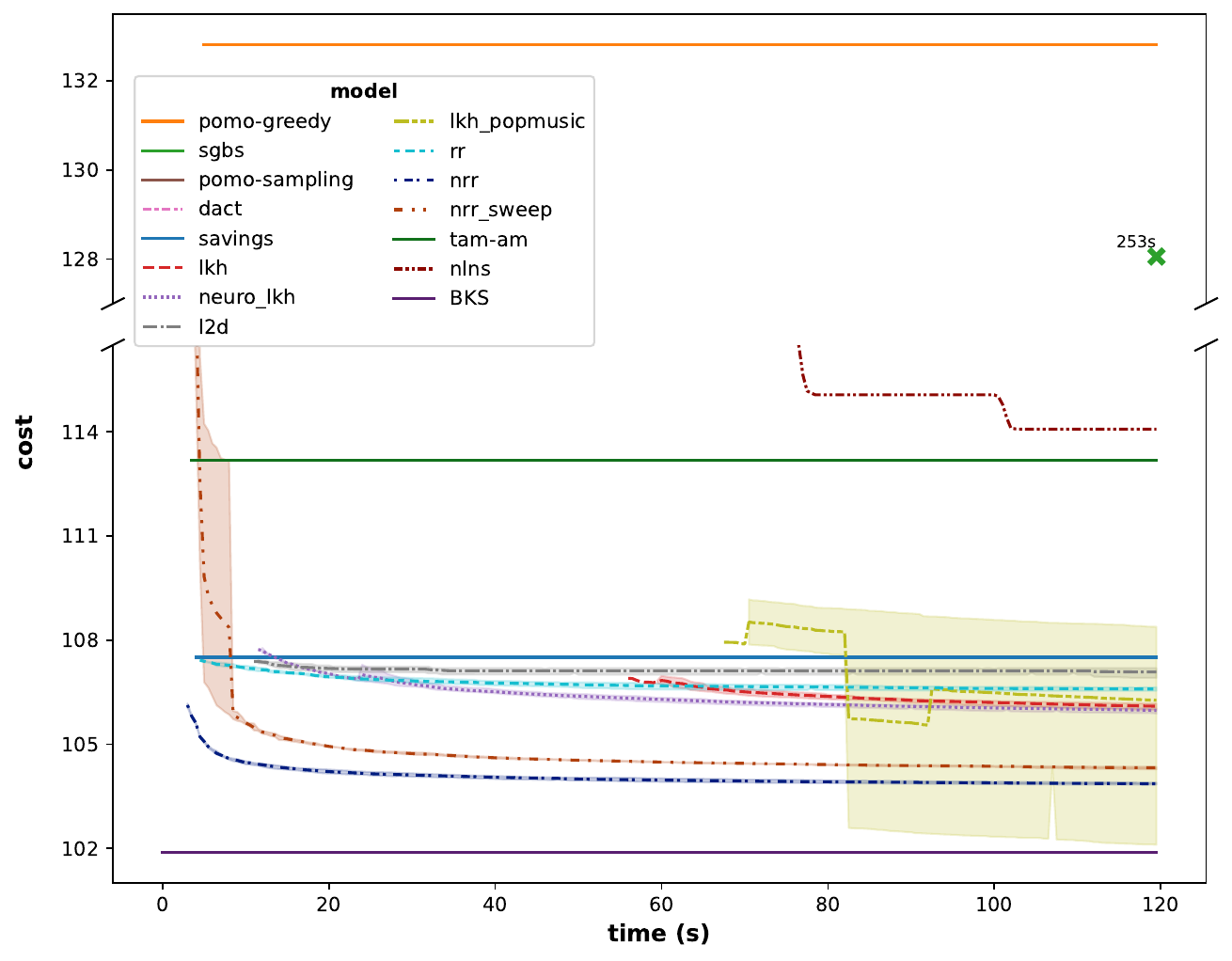}
		\captionof{figure}{
			Solution trajectories on all instances larger than $N=250$ of the Uchoa benchmark dataset~\cite{uchoa2017new} (total of 68 instances).
		}
		\label{fig:res_uchoa}
	\end{minipage}
	\hspace{3pt}
	\begin{minipage}{.49\textwidth}
		\centering
		\includegraphics[width=.999\linewidth]{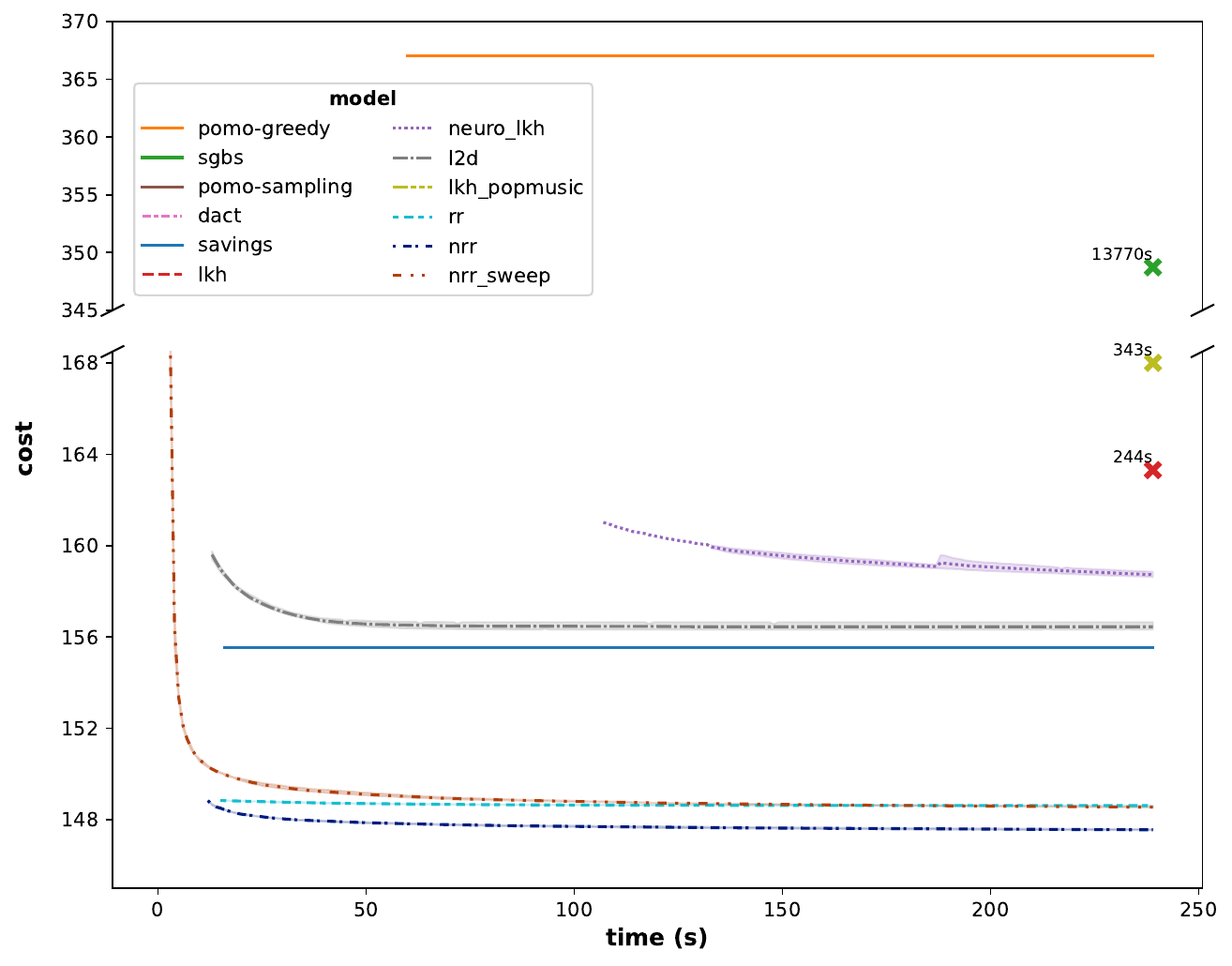}
		\captionof{figure}{
			Solution trajectories on the original real-world problems which were introduced in~\cite{li2021learning} (total of 50 instances).
		}
		\label{fig:res_real-l2d}
	\end{minipage}
\end{figure}

\section{Experiments}\label{s:experiments}
\textbf{Datasets} \ 
To evaluate the generalization performance of different methods we focus our comparison on the well-known Uchoa benchmark dataset~\cite{uchoa2017new}.
Moreover, we evaluate the models for which code is available on the original real world instances used in \cite{li2021learning} and our own dataset which consists of mixed uniform and clustered coordinate data of size $N \in [500, 1000, 2000, 4000 ]$ (see Appendix).

\textbf{Baselines} \ 
We compare our method against the state-of-the-art NC methods POMO~\cite{kwon2020pomo}, SGBS~\cite{choo2022simulation} and TAM-AM~\cite{hou2023generalize} as well as the neural improvement approaches NLNS~\cite{hottung2019neural}, L2D~\cite{li2021learning}, DACT~\cite{ma2021learning} and NeuroLKH~\cite{xin2021neurolkh}. Furthermore, we include the heuristic methods LKH3~\cite{helsgaun2000effective}, LKH-POP~\cite{helsgaun2018using,taillard19popmusic}, which uses POPMUSIC in combination with LKH3 and an implementation of the original RR procedure~\cite{schrimpf2000record} using random region selection and best insertion~\cite{mole1976sequential} as recreate step (cp. Alg. \ref{alg:rr}).

\textbf{Evaluation protocol} \ 
We run all methods for a fixed time budget of T = 60/120/240/480s for problems of size N = 500/1000/2000/4000 respectively. All hyperparameters as well as the used hardware are described in detail in the Appendix. In order to support the reproducibility of our results we will make our datasets and the code for our model and all experiments available with the publication.

\begin{table*}[th]
	\caption{
		Final cost of all methods on different datasets. Best result is shown in \textbf{bold}.
		Standard deviation over three runs  with different random seeds in brackets.
		For values marked with "*" the first results were only achieved (long) after the total time budget was exceeded (sometimes by an order of magnitude). 
		The DACT code breaks for $N=4000$ because of exp.\ increasing complexity which is why we report "NA".
		The results for TAM-AM and NLNS are only available for the Uchoa benchmark, thus we only report them in an extended table in the Appendix but show them in Fig.~\ref{fig:res_uchoa}. 
	}
	\label{tab:results_cvrp}
	\centering
	\tablesize
	\begin{tabular}{l|rrrrrr|r}
		\toprule
		\textbf{Model} & \textbf{N=500} & \textbf{N=1000} & \textbf{N=2000} & \textbf{N=4000} & \textbf{Uchoa} & \textbf{real world} & \textbf{Avg} \\
		\midrule
		\textbf{Savings}	&	39.7	(0.0) &	72.6	(0.0) &	139.8	(0.0) &	229.0	(0.0) &	107.5	(0.0) &	155.5	(0.0) &	124.0	\\
		\textbf{POMO} (g)	&	45.0	(0.0) &	112.4	(0.0) &	293.7	(0.0) &	575.8	(0.0) &	132.8	(0.0) &	367.0	(0.0) &	254.4	\\
		\textbf{POMO} (s)	&	75.1	(0.4) &	195.7	(0.9) &	454.8	(1.1) &	882.6	(1.9) &	205.0	(0.7) &	523.1	(1.4) &	389.4	\\
		\textbf{SGBS}		&	\textit{43.3}*	(0.0) &	\textit{105.9}*	(0.0) &	\textit{274.9}*	(0.0) &	\textit{539.4}*	(0.0) &	\textit{128.1}*	(0.0) &	\textit{348.7}*	(0.0) &	240.1	\\
		\textbf{LKH3}		&	\underline{38.4}	(0.1) &	\underline{71.3}	(0.2) &	141.1	(0.7) &	\textit{250.6}*	(3.7) &	106.0	(0.3) &	\textit{163.3}*	(2.5) &	128.5	\\
		\textbf{NeuroLKH}	&	\textbf{38.3}	(0.1) &	\textbf{71.1}	(0.2) &	140.3	(0.5) &	\textit{224.2}*	(44.7) &	105.9	(0.2) &	158.6	(0.5) &	123.1	\\  								
		\textbf{LKH-POP}	&	39.1	(0.2) &	73.6	(0.5) &	\textit{148.4}*	(1.1) &	244.4*	(5.7) &	\textbf{102.1}	(1.9) &	\textit{168.0}*	(1.6) &	129.3	\\ 
		\textbf{L2D}		&	39.3	(0.3) &	72.4	(0.4) &	141.1	(0.5) &	234.1	(1.0) &	106.9	(0.4) &	156.3	(0.5) &	125.0	\\
		\textbf{RR}			&	39.5	(0.0) &	72.4	(0.0) &	\underline{139.5}	(0.1) &	\underline{228.7}	(0.1) &	106.5	(0.1) &	148.6	(0.0) &	122.6	\\
		\textbf{DACT}		&	47.3	(0.0) &	83.6	(0.0) &	158.5	(0.0) &	NA			  &	122.5	(0.0) &	172.3	(0.0) &	139.4	\\
		\midrule
		\textbf{NRR}-sweep	&	39.0	(0.3) 		&	72.4	(0.5) &	141.6	(0.7) 	&	236.4	(0.7) 	&	104.3	(0.5) &	\underline{148.5}	(0.3) &	123.7	\\
		\textbf{NRR}		&	38.6	(0.1) &	71.6	(0.1) &	\textbf{138.9}	(0.1) &	\textbf{228.6}	(0.1) &	\underline{103.8}	(0.2) &	\textbf{147.5}	(0.1) &	\textbf{121.5}	\\
		
		\bottomrule
	\end{tabular}
\end{table*}

\begin{table*}[th]
	\caption{
		Area under Savings curve (AUSC). The shown methods achieved at least once a value better than 1.1 $\times$ the savings cost. All other considered baselines always have an AUSC of 0.0.
	}
	\label{tab:ausc}
	\centering
	\tablesize
	\begin{tabular}{l|rrrrrrr}
		\toprule
		\textbf{Dataset} &	\textbf{LKH3}	&	\textbf{NeuroLKH}	&	\textbf{LKH-POP}	&	\textbf{L2D}	&	\textbf{RR}		&	\textbf{NRR-swp}	&	\textbf{NRR}	\\
		\midrule
		\bm{$N=500$}	&	0.0282	&	\textbf{0.0312}	&	0.0064	&	0.0060	&	0.0050	&	0.0131	&	0.0253	\\
		\bm{$N=1000$}	&	0.0105	&	0.0098			&	0.0		&	0.0003	&	0.0021				&	0.0010		&	\textbf{0.0122}	\\
		\bm{$N=2000$}	&	0.0		&	0.0				&	0.0		&	0.0		&	\textbf{0.0048}					&	0.0		&	0.0011	\\
		\bm{$N=4000$}	&	0.0		&	0.0				&	0.0		&	0.0		&	0.0007							&	0.0		&	\textbf{0.0014}	\\
		\textbf{Uchoa}	&	0.0056	&	0.0094			&	0.0033	&	0.0032	&	0.0067	&	0.0254	&	\textbf{0.0313}	\\
		\textbf{real world}	&	0.0	&	0.0				&	0.0		&	0.0		&	0.0415	&	0.0418	&	\textbf{0.0477}	\\
		\bottomrule
	\end{tabular}
\end{table*}

\textbf{Results} \ 
We report the final cost (total length of all tours)
in Table~\ref{tab:results_cvrp} and plot the solution trajectories for the Uchoa and real world dataset in Fig.~\ref{fig:res_uchoa} and~\ref{fig:res_real-l2d}. NRR shows very strong performance, in particular for larger and more difficult instances. It outperforms all NC methods as well as most improvement approaches (L2D, RR and DACT) on all datasets. 
Only the very complex LKH methods (LKH3, NeuroLKH and LKH-POP) are in some cases able to slightly outperform our method. 
However, the large standard deviation (STD) for some of these methods on the $N=4000$ and Uchoa dataset demonstrate that these approaches do not reliably converge to such good results and often can lead to considerably worse performance than NRR. This can be seen in Fig.~\ref{fig:res_uchoa} where LKH-POP achieves a good result with one "lucky" seed while the other runs are far worse, leading to the large STD band displayed. The fluctuations in the mean cost also show that some runs require substantial run time to even find a first solution. In contrast, our method shows reliable performance with consistently low STD and significantly outperforms LKH-POP on all other datasets. Furthermore, many of the competitive results of baselines were only achieved after vastly exceeding the time budget (marked with "*" in Table~\ref{tab:results_cvrp} and displayed with an "\textbf{x}" marker labeled with the actual run time in the plots). 
Our comparison of the original NRR algorithm using Savings initialization with the alternative Sweep init shows that our method is also capable to achieve competitive performance even when starting from very bad initial solutions. This can also be seen in Fig.~\ref{fig:res_uchoa} and \ref{fig:res_real-l2d} where \textit{nrr\_sweep} shows a very steep initial decrease in cost while some other methods are not even able to beat the Savings baseline (we want to stress here again that the Savings algorithm is a very simple heuristic method). Moreover, the figures show that the anytime performance of our method is particularly good. 
To verify this,
we compute the area between each solution trajectory and a baseline cost. Taking inspiration from the DIMACS challenge~\cite{dimacs22} we set this baseline cost to 110\% of the cost achieved by the savings algorithm, which is the fundamental method on which our analysis is based. Then we re-normalize the area by this cost to achieve values in [0, 1]. By analogy to the area under curve (AUC) metric we call this measure \textit{Area under Savings curve} (AUSC) and report the results in Table~\ref{tab:ausc} (the exact calculation is described in the Appendix). As can be seen, our method exhibits very high values of AUSC, significantly outperforming all other methods in all but two cases.
Furthermore, in order to investigate the effect of different decisions in our algorithm design, we perform an additional ablation study (because of space constraints we report the corresponding results in the Appendix).

\section{Conclusion}
This paper presents a well motivated new approach to enable NC methods, which were learned on small training instances, to scale to routing problems which are up to 40$\times$ larger. NRR achieves this by introducing a meta procedure based on the ruin-recreate principle which is able to leverage the learned information of the neural construction model while focusing the improvement on regions with high potential.
The comprehensive experiments show that NRR is able to significantly improve the performance and generalization capabilities of NC models, being able to substantially outperform all construction methods and even some advanced improvement approaches.

\renewcommand\thesection{\Alph{section}}
\setcounter{section}{0}
\section{Appendix}

\subsection{SG construction heuristics}\label{s:sg_constr}

As described in the main paper, we use the tours $r$ as sub-structures to simplify the selection of sub-graphs.
One straight forward way of selecting such tours is to specify a number $K$ of nearest tour neighbors (measured by the euclidean distance between their geometric centers) and to select each existing tour as reference and adding its $K$ nearest neighbors to form the sub-graph consisting of $K+1$ tours. 
The obvious downside to this approach is that depending on the size of the tours $|r|$, the size $N_g$ of the resulting sub-graph $g$ can be much smaller or much larger than $N_{train}$ (the size of the instances in the NC training set). While in this case we can still assume that $N_g << N$, it can very well lead to worse performance. To circumvent this problem we propose two alternative methods for the construction of $G$. 
First, instead of fixing $K$, we can add neighboring tours sequentially until the SG size is very close to the size of the training instances, i.e.\ $|N_{\text{train}}-N_g| < \epsilon$. 
The second approach follows the same concept but is motivated by the sweep construction method adding tours sequentially to the SG when their center $\mu_r$ is passed by a beam which is rotated around the depot. Then, when $|N_{\text{train}}-N_g| < \epsilon$, the SG is complete and the next tour in sequence is added to a new SG.
The direct advantage of this method is that the resulting SGs are completely disjoint. This enables our method to better use the computation capacity of modern GPUs by solving a full batch of SGs via the NC method in parallel. 
Otherwise, if no disjoint SGs are required, then in order to create a larger and more diverse set of potential SGs we can rotate the beam several times, restarting it at different tour centers and afterwards remove any duplicates.

\subsection{AUSC}

\begin{figure}[h!]
	\centering
	\includegraphics[width=0.7\textwidth]{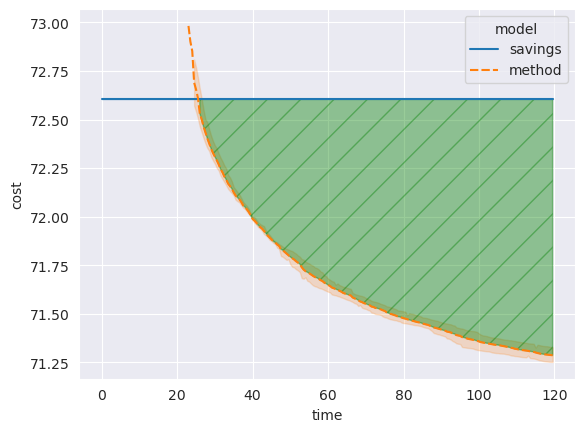}
	\caption{
		Visualization of AUSC. The green area between the savings cost and the method solution trajectory is the AUSC value.
	} 
	\label{fig:ausc}
\end{figure}

We propose the \textit{Area under Savings curve} (AUSC) as metric to measure the anytime performance of the compared methods w.r.t.\ a baseline cost, in our case the cost achieved by the savings algorithm. Since many methods are not even able to beat this very simple heuristic, we set the actual baseline cost to 110\% of the savings performance. Then we compute the full area below the savings curve and the solution cost trajectory of each method and normalize it by the total area under the savings curve:

\begin{equation}
	\text{AUSC}(c_{\text{m}}) = \frac{\int_{0}^{T}c_{\text{savings}} - \int_{0}^{T}\min(c_{\text{m}}, c_{\text{savings}})}{\int_{0}^{T}c_{\text{savings}}}
\end{equation}
where $T$ is the total time budget for the solution and $c_{\text{m}}$ is the cost trajectory of the corresponding method \textbf{m} while $c_{\text{savings}}$ is $1.1 \times$ the constant savings cost. In order to compute the area under curves with discrete time stamps of measurements, we use the composite trapezoidal rule. We visualize the concept in Figure~\ref{fig:ausc}

\subsection{Dataset generation}
\begin{figure}%
	\centering
	\subfloat{{\includegraphics[width=0.23\textwidth]{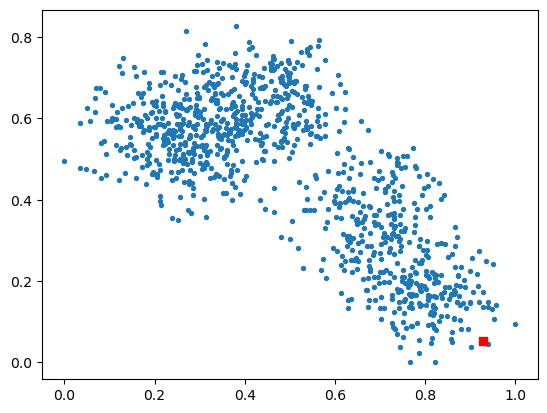} }}%
	\
	\subfloat{{\includegraphics[width=0.23\textwidth]{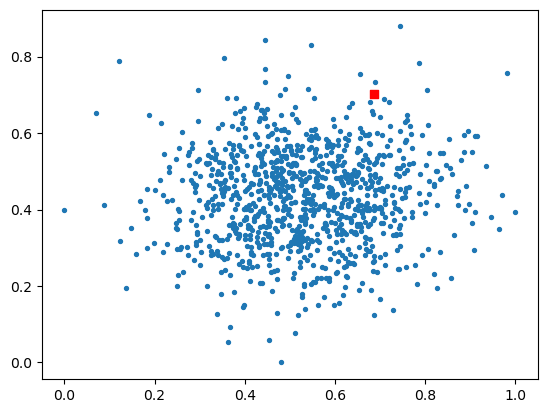} }}%
	\
	\subfloat{{\includegraphics[width=0.23\textwidth]{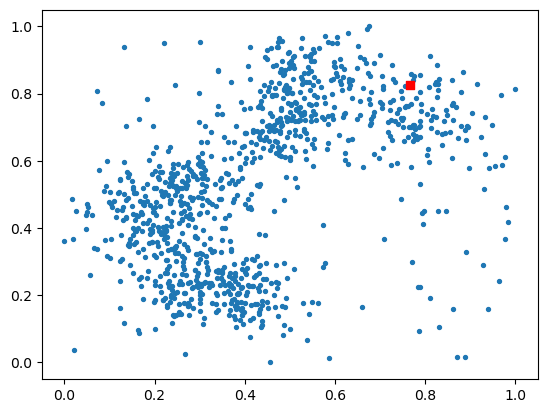} }}%
	\
	\subfloat{{\includegraphics[width=0.23\textwidth]{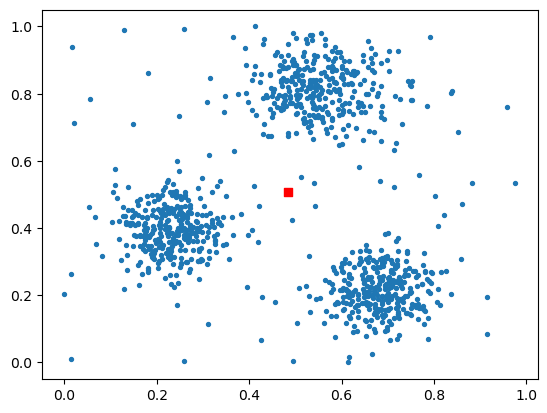} }}%
	\\
	\vspace{-10pt}
	\subfloat{{\includegraphics[width=0.23\textwidth]{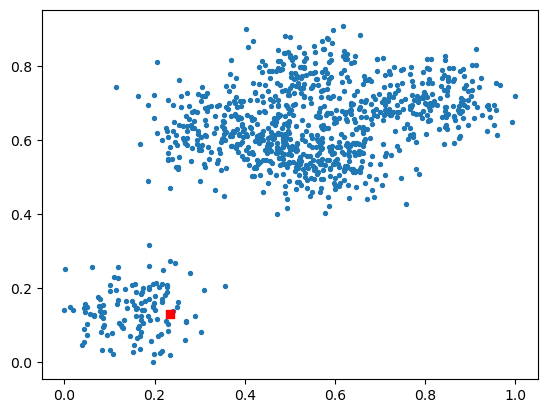} }}%
	\
	\subfloat{{\includegraphics[width=0.23\textwidth]{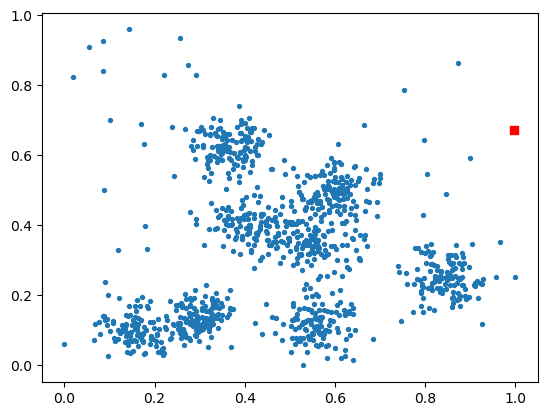} }}%
	\
	\subfloat{{\includegraphics[width=0.23\textwidth]{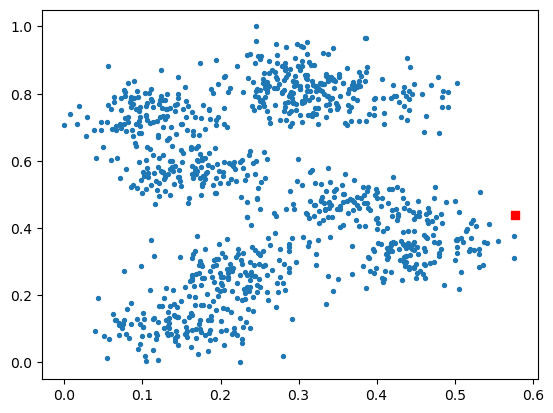} }}%
	\
	\subfloat{{\includegraphics[width=0.23\textwidth]{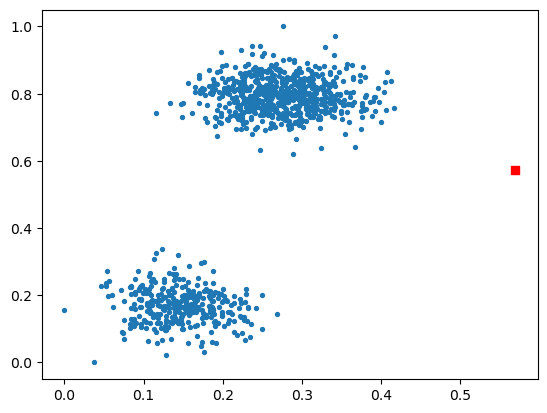} }}%
	\caption{
		Example plots showing the coordinates of generated instances. The red square represents the depot node.
	}%
	\label{fig:inst_coords}%
\end{figure}

We generate datasets with different numbers $N \in \{500, 1000, 2000, 4000 \}$ of customer nodes. The coordinates are sampled based on a mix of clustered and uniform data. 
For the uniform data we simply sample uniformly from the unit square like prior work~\cite{nazari2018reinforcement, kool2018attention}. 
The clustered data is sampled from a Gaussian Mixture Model where the number $K$ of mixture components is randomly selected between 1 and 10 for each instance. The mean of the components is sampled from a standard Normal distribution $\mu \sim N(0, 1)$ and the (diagonal) covariance matrix $\Sigma$ is sampled uniformly from $[0.05, 0.1]$. The weights are sampled as random integers between 1 and 9, normalized by a homogeneous constant vehicle capacity $Q = 50$ for all problem sizes. Finally, the fraction of uniformly sampled points compared to clustered points is sampled according to a beta distribution with parameters $\alpha = 0.5$ and $\beta = 9$. The resulting sampling procedure is able to generate diverse problem instances with varying coordinate distributions. In Figure \ref{fig:inst_coords} we show some examples of the resulting instances.

\subsection{Hyperparameters and hardware}

\paragraph{Neural Scoring Function}
The embedding dimension $d_{\text{emb}}$ of our neural scoring function $f_{\theta}$ is set to 128 as well as the hidden dimension of the node encoder and sub-graph encoder to 128, whereas the decoder uses a hidden dimension of 256.
The node encoder uses 4 GNN~\cite{morris2019weisfeiler} layers while the SG encoder and the decoder use 3 linear feed forward layers each. For each layer we utilize ReLU~\cite{nair2010rectified} activation functions and additional layer normalization~\cite{ba2016layer}. Between GNN layers we add residual connections. For the decoder we add dropout with a dropout probability of 0.1. As pooling operators for the node encoder we use summation and standard deviation while the SG encoder employs summation and maximum pooling, which are invariant with respect to padding the inputs with zeros in order to create SGs of similar size for batched processing. 
For training we use the Huber-loss~\cite{huber1992robust} as loss function and the Adam optimizer~\cite{kingma2014adam} with an initial learning rate of 0.0005, which is halved every 35 epochs. Moreover, we clip the gradient norm above a value of 0.5. We train our model for 80 epochs with a batchsize of 128. The KNN neighborhood graph $\G$ which is used for the node neighborhoods $\mathcal{H}$ is created with $K = 25$.

\paragraph{Neural Ruin Recreate}
For the training of the NC method employed in NRR we use the original hyperparameters reported in~\cite{kwon2020pomo, choo2022simulation}. The experiments on uniform data are performed with the checkpoints provided by the authors. For the other experiments we retrain the POMO model (which is also used for the POMO and SGBS baselines) on mixed data of size $N=100$.
NRR is initialized with a savings solution. It uses the sweeping based tour selection for the SG creation (see section~\ref{s:sg_constr}) to create disjoint sub-graphs. Then it scores these SGs via $f_{\theta}$ and selects up to 16 SGs which are fed to the NC method to be reconstructed. As NC method we employ the POMO~\cite{kwon2020pomo} method with sampling and a sample size of 1024. For acceptance we use simulated annealing (SA)~\cite{kirkpatrick1983optimization} with restarts after 25 iterations without improvement.

All models and baselines were run on a Intel Xeon Gold 6230 CPU (with 8 cores available) and if required a NVIDIA GeForce RTX 3090 GPU.
The model and experiment code as well as the used datasets are released via github: \url{https://github.com/jokofa/NRR}.

\section{Ablation Studies}
\subsection{Neural scoring function}
In order to evaluate the effect of our design decisions for the neural scoring function we report the results of an ablation study in table~\ref{tab:nsf_abl}.

\begin{table*}[h!]
	\caption{
		Ablation study for neural scoring function model run on the validation set of scored sub-graphs. We report mean absolute error (MAE) and mean squared error (MSE).
	}
	\label{tab:nsf_abl}
	\centering
	\tablesize
	\begin{tabular}{l|rr}
		\toprule
		\textbf{Configuration} &	\textbf{MAE}	&	\textbf{MSE} \\
		\midrule
		\textit{original} 	& 0.1045	& 	0.0239 	\\
		MSE loss 	& 0.1091	& 	0.0263	\\
		GAT			& 0.1176 	&	0.0328  \\
		max pool	& 0.1056 	&	0.0249 	\\
		mean pool	& 0.1053	&	0.0253 	\\
		sum pool	& 0.1048	&	0.0261 	\\
		no aggr		& 0.1050	&	0.0254	\\						
		\bottomrule
	\end{tabular}
\end{table*}

We compare our original model using HuberLoss, GraphConv~\cite{morris2019weisfeiler} sum and std pooling and aggregation of the node embeddings over all GNN layers with MSELoss, graph attention networks (GAT)~\cite{velivckovic2017graph}, different pooling operators and no aggregation. 
The results show that the HuberLoss works much better than MSE loss and that the simple GNN proposed in~\cite{morris2019weisfeiler} significantly outperforms the much more advanced GAT.
Although the effect of different pooling operators and the aggregation is smaller, our configuration shows the best results, in particular in terms of MSE.

\subsection{Neural Ruin Recreate (NRR)}
A second ablation study is performed to evaluate the effect of different configurations for the NRR procedure. All models are run on the mixed data validation set of size $N=500$. We compare different construction methods (see section~\ref{s:sg_constr}), selection approaches and different solver modes for the NC method, sampling (with the corresponding number of samples) vs. greedy (which always uses a rollout for each customer node, i.e.\ $N$). Furthermore, we compare SA to greedy acceptance of improving moves.

\begin{table*}[h!]
	\caption{
		Ablation study for NRR. We report the final cost, i.e.\ total length of all routes.
	}
	\label{tab:nrr_abl}
	\centering
	\tablesize
	\begin{tabular}{lllllll|r}
		\toprule
		\textbf{init} &	\textbf{construct} & \textbf{select} & \textbf{n\_mult} & \textbf{mode} & \textbf{n\_samp} &	\textbf{accpt} &	\textbf{cost} \\
		\midrule
		\textit{savings}	&	\textit{sweep}	&	\textit{disjoint}	&	\textit{16}	&	\textit{sampl}	&		\textit{1024}	&	\textit{SA}	&	\textbf{36.964}	\\
		\hline
		savings	&	sweep	&	disjoint	&	12	&	sampl	&		1024	&	SA	&	37.005	\\
		savings	&	sweep	&	disjoint	&	8	&	sampl	&		1024	&	SA	&	37.015	\\
		savings	&	sweep	&	disjoint	&	24	&	sampl	&		1024	&	SA	&	37.015	\\
		savings	&	sweep	&	disjoint	&	32	&	sampl	&		1024	&	SA	&	36.970	\\
		savings	&	sweep	&	disjoint	&	8	&	sampl	&		2048	&	SA	&	37.042	\\
		savings	&	sweep	&	disjoint	&	24	&	sampl	&		2048	&	SA	&	37.052	\\
		savings	&	sweep	&	disjoint	&	32	&	sampl	&		2048	&	SA	&	36.981	\\
		savings	&	sweep	&	disjoint	&	8	&	sampl	&		512		&	SA	&	36.992	\\
		savings	&	sweep	&	disjoint	&	12	&	sampl	&		512		&	SA	&	36.997	\\
		savings	&	sweep	&	disjoint	&	16	&	sampl	&		1024	& 	greedy	&	37.013	\\
		savings	&	sweep	&	disjoint	&	12	&	sampl	&		1024	& 	greedy	&	36.993	\\
		savings	&	sweep	&	disjoint	&	8	&	sampl	&		1024	& 	greedy	&	36.993	\\
		savings	&	sweep	&	disjoint	&	8	&	sampl	&		2048	& 	greedy	&	36.996	\\
		\hline
		savings	&	sweep	&	greedy		&	1	&	greedy	&		$N$		&	SA	&	37.587	\\
		savings	&	sweep	&	greedy		&	1	&	sampl	&		512		&	SA	&	37.462	\\
		savings	&	sweep	&	multi		&	8	&	greedy	&		$N$		&	SA	&	37.226	\\
		savings	&	sweep	&	multi		&	8	&	sampl	&		512		&	SA	&	37.049	\\
		savings	&	knn	&	greedy		&	1	&	greedy	&		$N$		&	SA	&	37.448	\\
		savings	&	knn	&	greedy		&	1	&	sampl	&		512		&	SA	&	37.720	\\
		savings	&	knn	&	multi		&	8	&	greedy	&		$N$		&	SA	&	37.284	\\
		savings	&	knn	&	multi		&	8	&	sampl	&		512		&	SA	&	37.515	\\
		savings	&	add\_nn	&	greedy	&	1	&	greedy	&		$N$		&	SA	&	37.338	\\
		savings	&	add\_nn	&	greedy	&	1	&	sampl	&		512		&	SA	&	37.312	\\
		savings	&	add\_nn	&	multi	&	8	&	greedy	&		$N$		&	SA	&	37.033	\\
		savings	&	add\_nn	&	multi	&	8	&	sampl	&		512		&	SA	&	37.342	\\
		savings	&	add\_nn	&	multi	&	16	&	greedy	&		$N$		&	SA	&	37.063	\\
		savings	&	add\_nn	&	multi	&	16	&	sampl	&		128		&	SA	&	37.338	\\
		\bottomrule
	\end{tabular}
\end{table*}

The results in table~\ref{tab:nrr_abl} show the effect of the chosen sweep construction and disjoint SG selection methods for different numbers of selected SGs and samples. The first row represents our original model configuration. Under the time budget of 60s our configuration seems to be a good trade-off between the parallelization capacity on the GPU using more SGs and the solver performance controlled by the number of samples. SA which also accepts solutions which are slightly worse than the current best seems to help to escape from local optima which are encountered by greedy acceptance. Other combinations of construction and selection methods like sweep construction with greedy or multi selection as well as knn or add\_nn (which adds neighboring tours until the SG size is close to a specified value) lead to significantly worse performance.

\clearpage
\section{Additional plots}

\begin{figure}[h]
	\centering
	\includegraphics[width=0.999\textwidth]{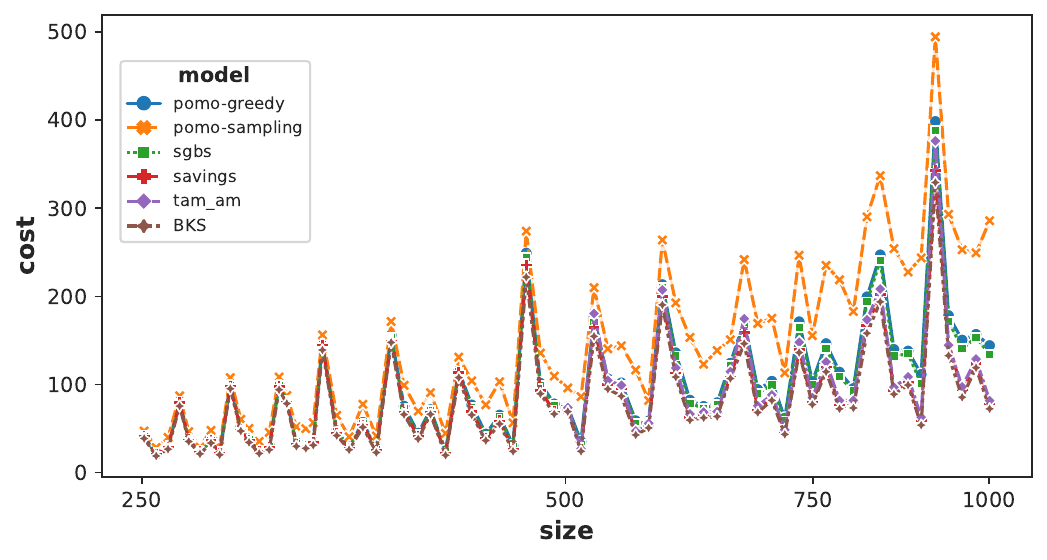}
	\caption{
		Results of constructive methods POMO~\cite{kwon2020pomo} (greedy, sampling), SGBS~\cite{choo2022simulation} and Clarke-Wright savingsings~\cite{clarke1964scheduling} on the Uchoa benchmark~\cite{uchoa2017new}. Non binned plots which show high fluctuation because of the different recurring configurations for the coordinate and demand distribution. Note that for TAM-AM only results for $N > 500$ were reported in \cite{hou2023generalize}. We also report the \emph{best known solution} (BKS).
	} 
	\label{fig:res_uchoa2}
\end{figure}

\begin{figure}[h]
	\centering
	\includegraphics[width=0.8\textwidth]{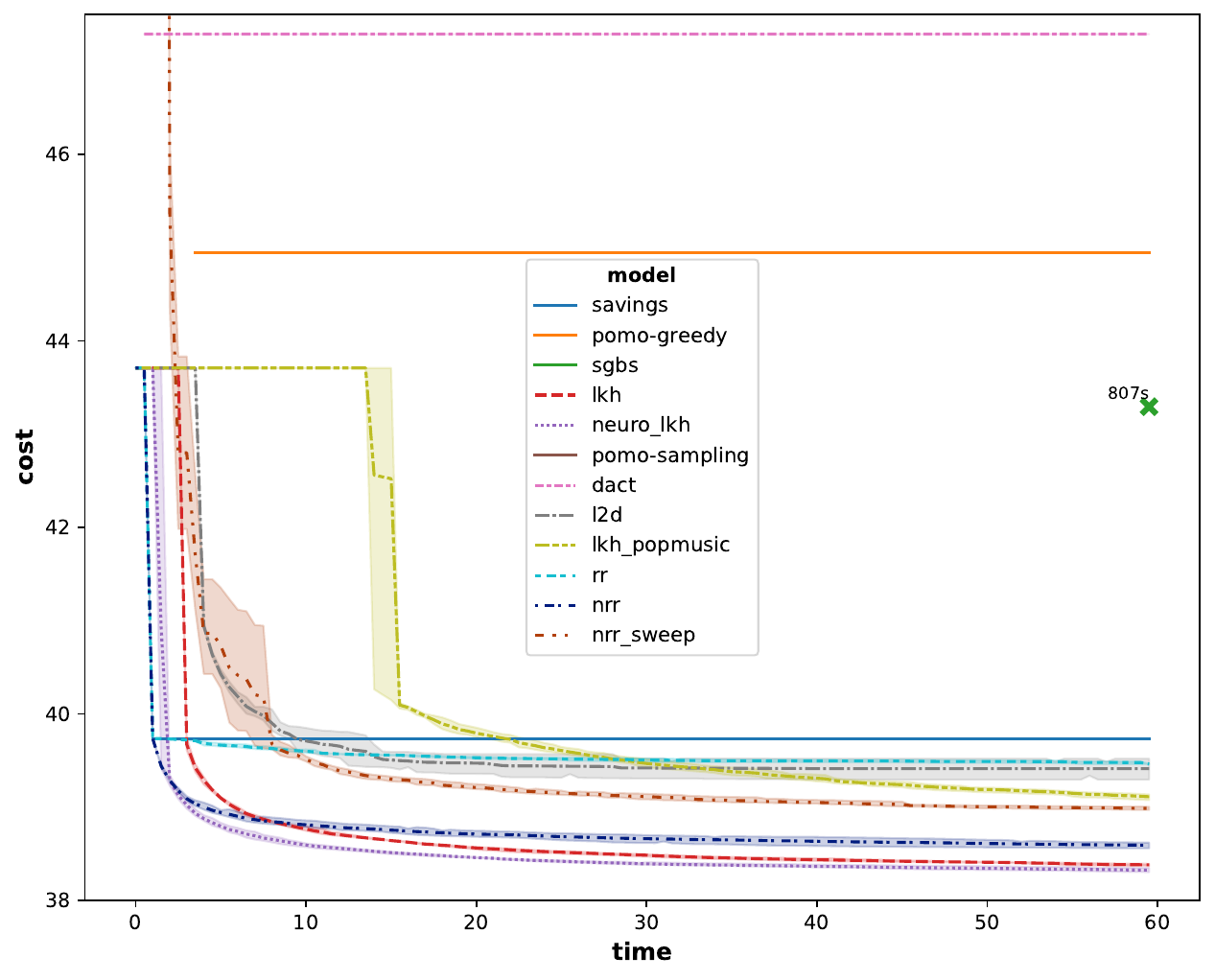}
	\caption{
				Solution trajectories for methods on 50 problem instances of size $N=500$ from a mixed uniform and clustered distribution.
				All methods were run for 3 random seeds, displaying the standard deviation bands. Methods which did not produce a first feasible result within the time budget are shown with a "x" marker on the right of the figure, labeled with the actual run time.
			} 
	\label{fig:res_500}
\end{figure}

\begin{figure}[h]
	\centering
	\includegraphics[width=0.8\textwidth]{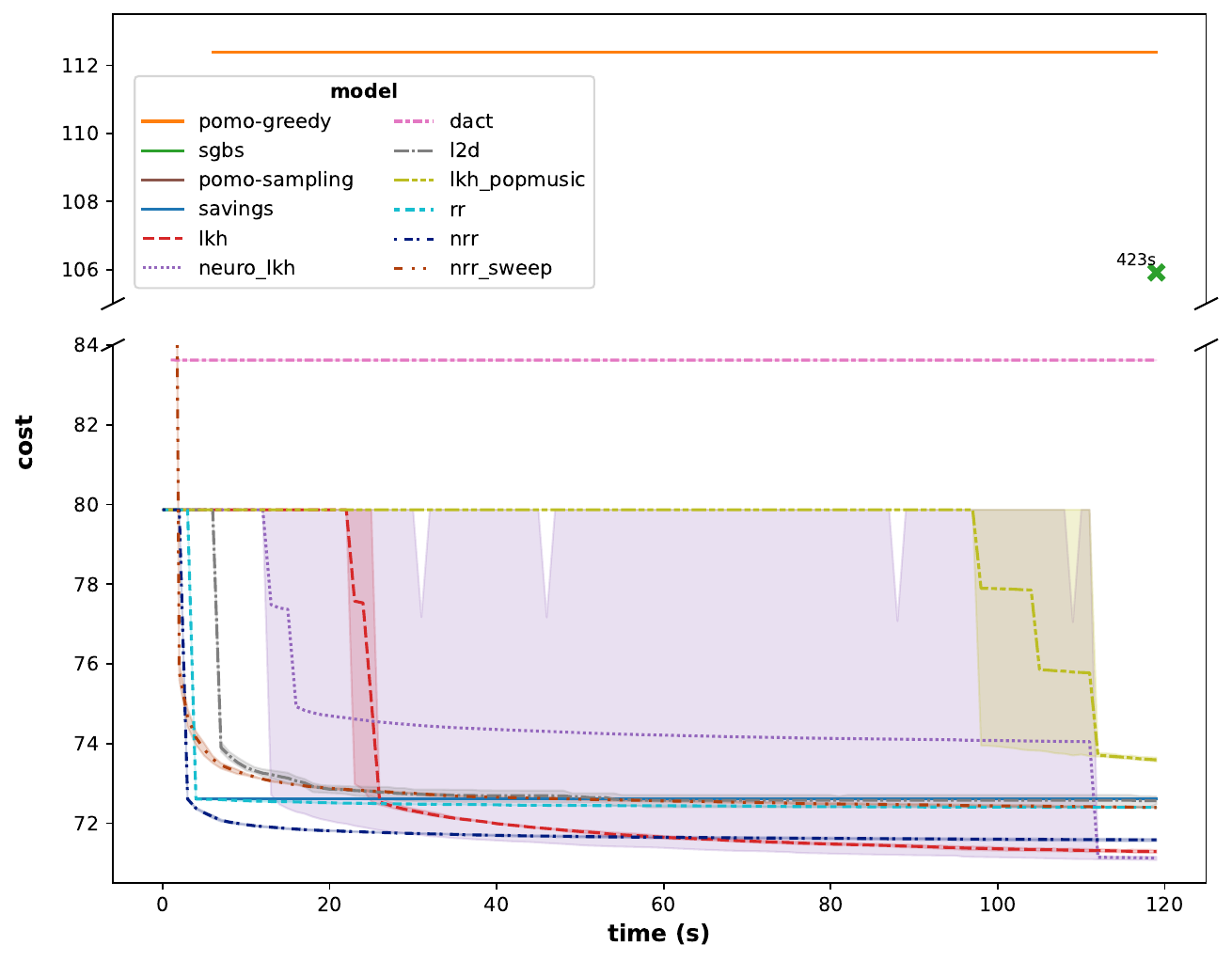}
	\caption{
		Solution trajectories for methods on 50 problem instances of size $N=1000$ from a mixed uniform and clustered distribution.
		All methods were run for 3 random seeds, displaying the standard deviation bands. Methods which did not produce a first feasible result within the time budget are shown with a "x" marker on the right of the figure, labeled with the actual run time.
	} 
	\label{fig:res_1000}
\end{figure}

\clearpage

\bibliographystyle{abbrv}
\bibliography{references}


\end{document}